Running title: visual processing of adversarial images in CNNs and the human brain

# Dissociable neural representations of adversarially perturbed images in convolutional neural networks and the human brain

Chi Zhang[1], Xiaohan Duan[1], Linyuan Wang[1], Yongli Li[2], Bin Yan[1], Guoen Hu[1], Ru-Yuan Zhang[3*], Li Tong[1*]

[1] National Digital Switching System Engineering and Technological Research Center, Zhengzhou, China 450000
[2] People's Hospital of Henan Province, Zhengzhou, China, 450000
[3] Shanghai Key Laboratory of Psychotic Disorders, Shanghai Mental Health Center, Shanghai Jiao Tong University School of Medicine, Shanghai, China. 200030
[4] Institute of Psychology and Behavioral Science, Shanghai Jiao Tong University, Shanghai, China. 200030
* co-senior authors

Ru-Yuan Zhang: ruyuanzhang@gmail.com. ORCID: 0000-0002-0654-715X

Corresponding authors:
Ru-Yuan Zhang
315 Zhongyuan building, 1954 Huashan Road
Shanghai Jiao Tong University, Xuhui District, Shanghai, China. 200030
ruyuanzhang@gmail.com
15674496106

Li Tong
National Digital Switching System Engineering and Technological Research Center
No.62, Kexue Road, Zhengzhou City, China, 450000
tttocean_tl@hotmail.com

**ABSTRACT**

Despite the remarkable similarities between convolutional neural networks (CNN) and the human brain, CNNs still fall behind humans in many visual tasks, indicating that there still exist considerable differences between the two systems. Here, we leverage adversarial noise (AN) and adversarial interference (AI) images to quantify the consistency between neural representations and perceptual outcomes in the two systems. Humans can successfully recognize AI images as the same categories as their corresponding regular images but perceive AN images as meaningless noise. In contrast, CNNs can recognize AN images similar as corresponding regular images but classify AI images into wrong categories with surprisingly high confidence. We use functional magnetic resonance imaging to measure brain activity evoked by regular and adversarial images in the human brain, and compare it to the activity of artificial neurons in a prototypical CNN—AlexNet. In the human brain, we find that the representational similarity between regular and adversarial images largely echoes their perceptual similarity in all early visual areas. In AlexNet, however, the neural representations of adversarial images are inconsistent with network outputs in all intermediate processing layers, providing no neural foundations for the similarities at the perceptual level. Furthermore, we show that voxel-encoding models trained on regular images can successfully generalize to the neural responses to AI images but not AN images. These remarkable differences between the human brain and AlexNet in representation-perception association suggest that future CNNs should emulate both behavior and the internal neural presentations of the human brain.

**Keywords**: adversarial images, convolutional neural network, human visual cortex, functional magnetic resonance imaging, representational similarity analysis, forward encoding model.

**INTRODUCTION**

The recent success of convolutional neural networks (CNNs) in many computer vision tasks inspire neuroscientists to consider them as a ubiquitous computational framework to understand biological vision [1, 2]. Indeed, a bulk of recent studies have demonstrated that visual features in CNNs can accurately predict many spatiotemporal characteristics of brain activity [3-10]. These findings reinforce the view that modern CNNs and the human brain share many key structural and functional substrates [11].

Despite the tremendous progress, current CNNs still fall short in several visual tasks. These disadvantages suggest that critical limitations still exist in the modern CNNs[12]. One potent example is adversarially perturbed images, a class of images that can successfully "fool" even the most state-of-the-art CNNs [13, 14]. Adversarial noise (AN) images (Fig. 1B) look like meaningless noise to humans but can be classified by CNNs into familiar object categories with surprisingly high confidence [13]. Adversarial interference (AI) images are generated by adding a small amount of special noise to regular images (Fig. 1C). The special noise looks minimal to humans but severely impairs CNNs' recognition performance[14]. Therefore, adversarial images present a compelling example of double-dissociation between CNNs and the human brain, because artificially created images can selectively alter perception in one system without significantly impacting the other one.

It remains unclear the neural mechanisms underlying the drastically different visual behavior between CNNs and the human brain with respect to adversarial images. In particular, why do the two systems receive similar stimulus inputs but generate distinct perceptual outcomes? In the human brain, it has been known that the neural representations in low-level visual areas mostly reflect stimulus attributes whereas the neural representations in high-level visual areas mostly reflect perceptual outcomes[15, 16]. For example, the neural representational similarity in human inferior temporal cortex is highly consistent with perceived object semantic similarity [17]. In other words, there exist a well-established representation-perception association in the human brain.

This processing hierarchy is also a key feature of modern CNNs. If the representational architecture in CNNs truly resembles the human brain, we should expect similar neural substrates supporting CNNs' "perception". For CNNs, AI images and regular images are more similar at the pixel level but yield different perceptual outcomes. By contrast, AN images and regular images are more similar at the "perceptual" level. We would expect that AI and regular images have more similar neural representations in low-level layers while AN and regular images have similar neural representations in high-level layers. In other words, there must exist at least one high-level representational layer that supports the same categorical perception of AN and regular images, similar to the representation-perception association in the human brain. However, as we will show later in this paper, we find no representational pattern that supports RE-AN perceptual similarity in all intermediate representation layers except the output layer.

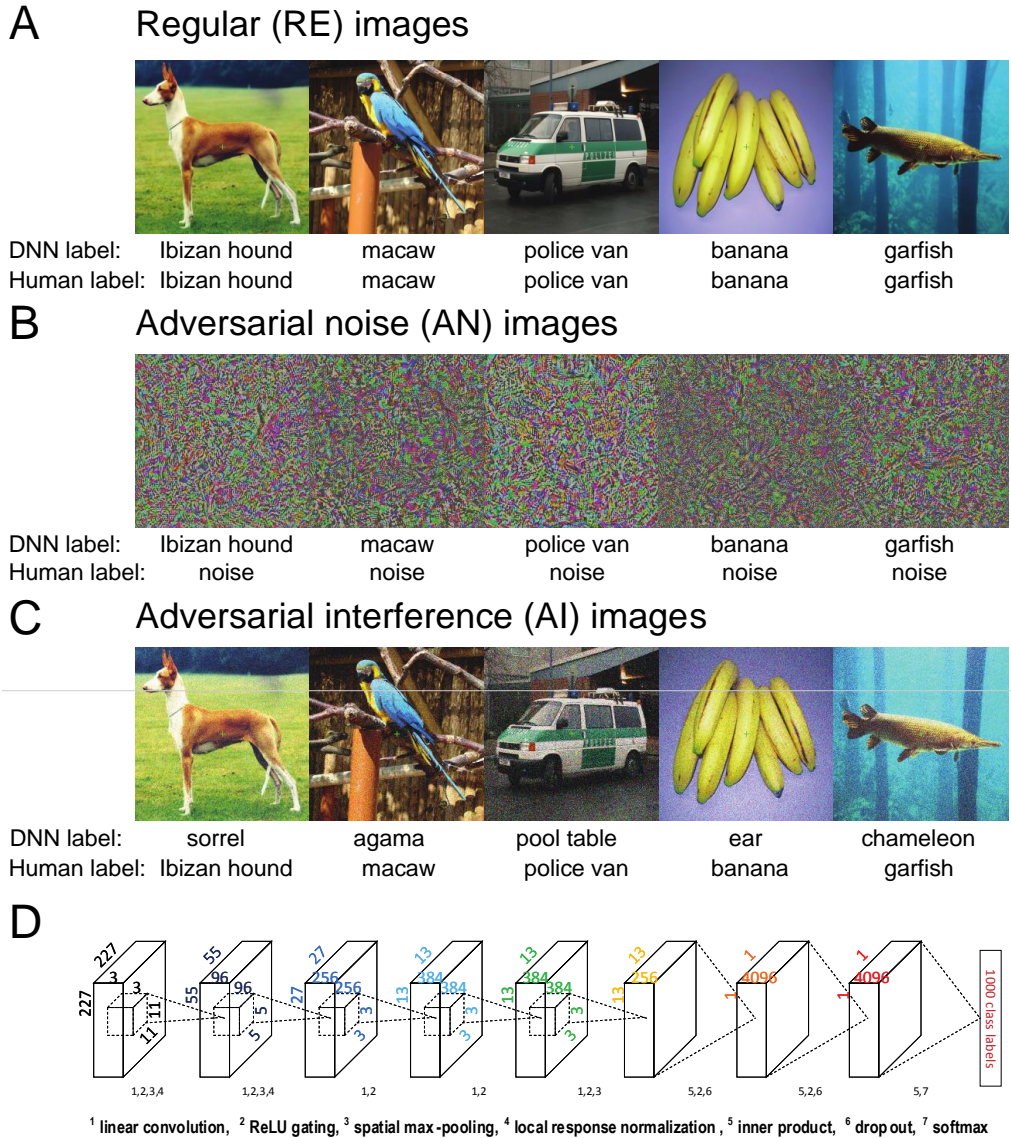

**Figure 1.** *A-C.* Example regular (RE, ***panel A***), adversarial noise (AN, ***panel B***) images and adversarial interference (AI, ***panel C***) images. The five AN and five AI images one-by-one correspond to the five RE images. The labels provided by AlexNet and humans are listed under the images. The AI images contain a small amount of special image noise but overall look similar to the corresponding RE images. Humans can easily recognize the AI images as corresponding categories but the AN images as noise AlexNet can classify the AN images into corresponding categories with over 99% confidence, but recognize the AI images as wrong categories. ***D.*** The architecture of AlexNet. Details have been documented in Krizhevsky A et al. [18]. Each layer uses some or all the following operations: linear convolution, ReLU gating, spatial max-pooling, local response normalization, inner product, dropout and softmax.

The majority of prior studies focused on revealing similarities between CNNs and the human brain. In this paper, we instead leverage adversarial images to examine the differences between the two systems. We particularly emphasize that delineating the differences here does not mean to object CNNs as a useful computational framework for human vision. On the contrary, we acknowledge the promising utilities of CNNs in modeling biological vision but we believe it is more valuable to understand differences rather than similarities such that we are in a better position to eliminate these discrepancies and construct truly brain-like machines. In this study, we use a well-established CNN—AlexNet and investigate the activity of artificial neurons towards adversarial images and their corresponding regular images. We also use functional magnetic resonance imaging (fMRI) to measure the cortical responses evoked by RE and adversarial images in humans. Representational similarity analysis (RSA) and forward encoding modeling allow us to directly contrast representational geometries within and across systems to understand the capacity and limit of both systems.

## MATERIALS AND METHODS

**Ethics statement.** All experimental protocols were approved by the Ethics Committee of the Henan Provincial People's Hospital. All research was performed in accordance with relevant guidelines and regulations. Informed written consent was obtained from all participants.

**Subjects.** Three healthy volunteers (one female and two males, aged 22~28 years) participated in the study. The subject S3 was the author C.Z. The other two subjects were naïve to the purpose of this study. All subjects were monolingual native-Chinese speakers and right-handed. All subjects had normal or corrected-to-normal vision and considerable experience of fMRI experiments.

**Convolutional Neural Network**. We chose AlexNet and implemented it using the Caffe deep learning framework [18] [19]. AlexNet consists of five convolutional layers and three fully-connected layers (Fig. 1D). The five convolutional layers each have 96, 256, 384, 384, and 256 linear convolutional kernels. The three fully-connected layers each have 4096, 4096, and 1000 artificial neurons. All convolutional layers perform linear convolution and rectified linear unit (ReLU) gating. Spatial max pooling is used only in layers 1, 2, and 5 to promote spatial invariance of sensory inputs. In layers 1 and 2, response normalization is implemented to simulate activity competition between local neurons. The ReLU activation function and dropout are used in fully-connected layers 6 and 7. Layer 8 uses the softmax function to output the probabilities for 1000 target categories. In our study, all images were resized to $227 \times 227$ pixels in all three RGB color channels.

**Image stimuli**

*Regular images*. Regular (RE) images in our study were sampled from the ImageNet database [19]. ImageNet is currently the most advanced benchmark database on which almost all state-of-the-art CNNs are trained for image classification. We selected one image (width and height > 227 pixels and aspect ratio > 2/3 and < 1.5) from each of 40 representative object categories. AlexNet can classify all images into their corresponding categories with probabilities greater than 0.99. The 40 images can be evenly divided into 5 classes: dogs, birds, cars, fruits, and aquatic animals.

*Adversarial images*. Adversarial images include adversarial noise (AN) (Fig. 1B) and adversarial interference (AI) images (Fig. 1C). A pair of AN and AI images were generated for each RE image. As such, a total of 120 images (40 RE + 40 AN + 40 AI) were used in the entire experiment.

The method to generate AN images has been documented in Nguyen A et al. [13]. We briefly summarize the method here. We first used the averaged image of all images in ImageNet as the initial AN image. Note that the category label of the corresponding RE image was known, and AlexNet had been fully trained. As such, we first fed the initial AN image to AlexNet and forwardly computed the probability for the correct category. This probability was expected to be initially low. We then used the backpropagation method to transduce error signals from the top layer to image pixel space. Pixel values in the initial AN image were then adjusted accordingly to enhance the classification probability. This process of forward calculation and backpropagation was iterated many times until the AN image converged.

We also included an additional regularization item to control the overall intensity of the image. Formally, let $P_c(I)$ be the probability of the class $c$ given an image $I$. We would like to find an $L_2$-regularized image $I^*$, such that it maximizes the following objective:

$$I^* = \arg\max_I P_c(I) - \lambda \left\| I - I_{mean} \right\|_2^2 , \qquad (1)$$

where $\lambda$ is the regularization parameter and $I_{mean}$ is the grand average of all images in ImageNet. This process was iterated until $P_c(I)$ reached 0.99. Note that the internal structure (i.e., all connection weights) of AlexNet was fixed throughout the entire training process, and we only adjusted pixel values in input AN images.

AI images were generated using a similar approach. We initialized an AI image as its corresponding RE image, and we then adjusted pixel values in the initial AI image to enhance the probability that the AI image was categorized into another class. All AI images were classified into a "wrong" class with probabilities greater than 0.99.

**Apparatus.** All computer-controlled stimuli were programmed in Eprime 2.0 and presented using a Sinorad LCD projector (resolution 1920 × 1080 at 120 Hz; size 89 cm × 50 cm; viewing distance 168 cm). Stimuli were projected onto a rear-projection monitor located over the head. Subjects viewed the monitor via a mirror mounted on the head coil. Behavioral responses were recorded by a button box.

**fMRI experiments**

*Main experiment*. Each subject underwent two scanning sessions in the main experiment. In each session, half of all images (20 images x 3 RE/AN/AI = 60 images) were presented. Each session consisted of five scanning runs, and each run contained 129 trials (2 trials per image and 9 blank trials). The image presentation order was randomized within a run. In a trial, a blank lasted 2 s and was followed by an image (12° × 12°) of 2 s. A 20 s blank period was included to the beginning and the end of each run to establish a good baseline and compensate for the initial insatiability of the magnetic field. A fixation point (0.2° × 0.2°) was shown at center-of-gaze, and participants were instructed to maintain steady fixation throughout a run. Participants pressed buttons to perform an animal judgment task—whether an image belongs to animals. The task aimed to engage subjects' attention onto the stimuli.

*Retinotopic mapping and functional localizer experiments*. A retinotopic mapping experiment was also performed to define early visual areas, as well as two functional localizer experiments to define lateral occipital (LO) lobe and human middle temporal lobe (hMT+)

The retinotopic experiment used standard phase-encoding methods [20]. Rotating wedges and expanding rings were filled by textures of objects, faces, and words, and were presented on top of achromatic pink-noise backgrounds (http://kendrickkay.net/analyzePRF/). Early visual areas (V1–V4) were defined on the spherical cortical surfaces of individual subjects.

The two localizer experiments were used to create a more precise LO mask (see region-of-interests definition section below). Each localizer experiment contained two runs. In the LO localizer experiment, each run consisted of 16 stimulus blocks and 5 blank blocks. Each run began with a blank block, and a blank block appeared after every 4 stimulus blocks. Each block lasted 16s. Intact images and their corresponding scrambled images were alternately presented in a stimulus block. Each stimulus block contained 40 images (i.e., 20 intact + 20 scramble images). Each image (12° × 12°) lasted 0.3 s and was followed by a 0.5 s blank.

In the hMT+ localizer experiment, each run contained 10 stimulus blocks, and each block lasted 32 s. In a block, a static dot stimulus (24 s) and a moving-dot stimulus (8 s) were alternately presented. All motion stimuli subtended a 12° × 12° square area on a black background. An 8 s blank was added to the beginning and the end of each run.

**MRI data acquisition**. All MRI data were collected using 3.0-Tesla Siemens MAGNETOM Prisma scanner and a 32-channel head coil at the Department of Radiology at the People's Hospital of Henan Province.

An interleaved T2*-weighted, single-shot, gradient-echo echo-planar imaging (EPI) sequence was used to acquire functional data (60 slices, slice thickness 2 mm, slice gap 0 mm, field of view 192 × 192 mm$^2$, phase-encode direction anterior-posterior, matrix size 96 × 96, *TR/TE* 2000/29 ms, flip angle 76°, nominal spatial resolution 2 × 2 × 2 mm$^3$). Three B0 fieldmaps were acquired to aid post-hoc correction for EPI spatial distortion in each session (resolution 2 × 2 × 2 mm$^3$, *TE$_1$* 4.92 ms, *TE$_2$* 7.38 ms, *TA* 2.2 min). In addition, high-resolution T1-weighted anatomical images were also acquired using a 3D-MPRAGE sequence (*TR* 2300 ms, *TE* 2.26 ms, *TI* 900 ms, flip angle 8°, field of view 256 × 256 mm$^2$, voxel size 1. × 1. × 1. mm$^3$).

**MRI data preprocessing**. The pial and the white surfaces of subjects were constructed using FreeSurfer softerware (http://surfer.nmr.mgh.harvard.edu). An intermediate gray matter surface between the pial and the white surfaces was also created for each subject.

For functional data, we discarded the data points of the first 18 s in the main experiment, the first 14 s in the LO localizer experiment, and the first 6 s in the hMT+ localizer experiment. Functional data then underwent slice time correction using cubic interpolation. The regularized time-interpolated field maps were used to correct EPI spatial distortion. Rigid-body motion parameters were then estimated from the undistorted EPI volumes. Finally, the effects of distortion correction, head motion correction, and data mapping from volumes to surfaces were concatenated and performed using a single interpolation step to maximally preserve spatial resolution.

**General linear modeling**. We estimated the vertex responses (i.e., beta estimates from GLM modeling) of all stimulus trials in the main experiment using the GLMdenoise method [21].

All blank trials were modeled as a single predictor. This analysis yielded beta estimations of 241 conditions (120 images × 2 trials + 1 blank trial). Notably, we treated two presentations of the same image as two distinct predictors in order to calculate the consistency of the response patterns across the two trials.

**Region-of-interest (ROI) definitions**. Based on the retinotopic experiment, we calculated the population receptive field (pRF) (http://kendrickkay.net/analyzePRF) of each vertex and defined low-level visual areas (V1–V4) based on the pRF maps. To define LO, we first selected vertices that show significantly higher responses to intact images than scrambled images (two-tails t-test, $p < 0.05$, uncorrected). In addition, hMT+ was defined as the area that shows significantly higher responses to moving than static dots (two-tails t-test, $P < 0.05$, uncorrected). The intersection vertices between LO and hMT+ were `then removed from LO.

**Vertex selection**. To further select task-related vertices in each ROI, we performed a searchlight analysis on flattened 2D cortical surfaces [22]. For each vertex, we defined a 2D searchlight disk with 3 mm radius. The geodesic distance between two vertices was approximated by the length of the shortest path between them on the flattened surface. Given the vertices in the disk, we calculated the representational dissimilarity matrices (RDM) of all RE images for each of the two presentation trials. The two RDMs were then compared (Spearman's R) to show the consistency of activity patterns across the two trials. The 200 vertices (100 vertices from each hemisphere) with the highest correlation values were selected in each ROI for further analysis (Fig. 3). Note that vertex selection was only based on the responses to the RE images and did not involve any response data for the AN and the AI images. We also selected a total of 400 vertices in each area and we found our results held. The results are shown in Fig. S2.

**Representational similarity analysis**. We applied RSA separately to the activity in the CNN and in the brain.

*RSA on CNN layers and brain ROIs*. For one CNN layer, we computed the representational dissimilarity between every pair of the RE images, yielding a 40 × 40 RDM (i.e., $RDM_{RE}$) for the RE images. Similarly, we obtained the other two RDMs each for the AN (i.e., $RDM_{AN}$) and the AI images (i.e., $RDM_{AI}$). We then calculated the similarity between the three RDMs as follows:

$$R_{RE-AN} = corr(RDM_{RE}, RDM_{AN}), \qquad (2)$$

$$R_{RE-AI} = corr(RDM_{RE}, RDM_{AI}), \qquad (3)$$

This calculation generated one RE-AN similarity value and one RE-AI similarity value for that CNN layer (see Fig. 2B). We repeated the same analysis above on the human brain except that we used the activity of vertices in a brain ROI.

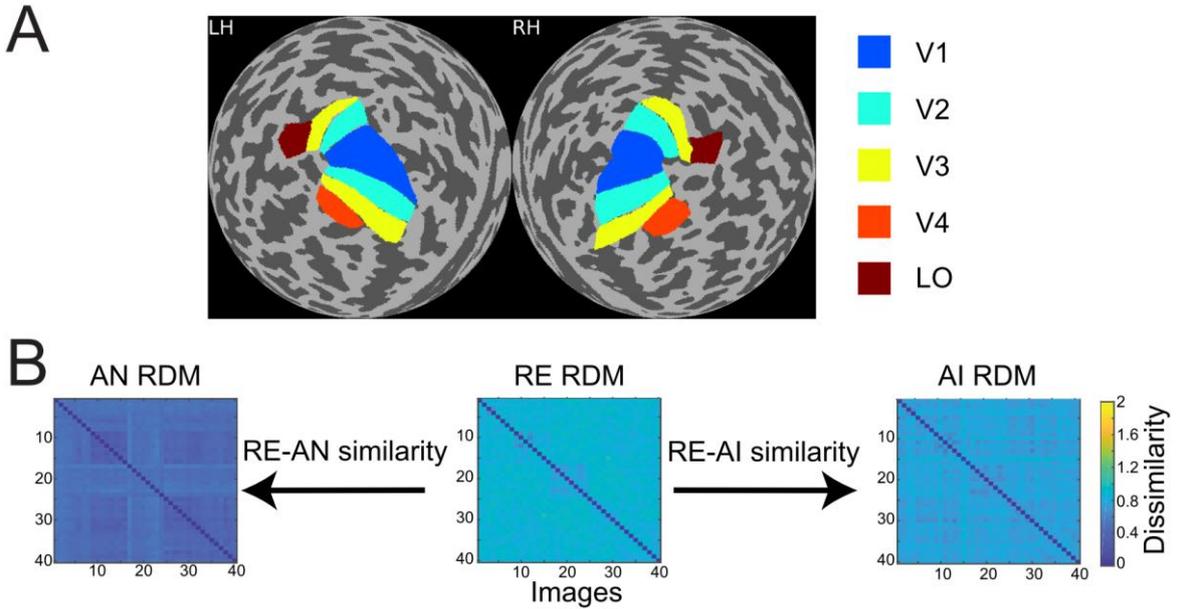

**Figure 2.** ***A.*** Regions of interest (ROIs) in a sample subject. Through retinotopic mapping and functional localizer experiments, we identified five ROIs—V1, V2, V3, V4 and lateral occipital (LO) cortex—in both left (LH) and right (RH) hemispheres. ***B.*** Calculation of RE-AN and RE-AI similarity. For each CNN layer or brain ROI, three RDMs are calculated with respect to the three types of images. We then calculate the Spearman correlation between the AN and the RE RDMs, obtaining the RE-AN similarity. Similarly, we can calculate the RE-AI similarity.

To examine whether the RE-AN or the RE-AI similarity values significantly deviated from null hypotheses, we randomized the image labels with respect to the endowed activity 1000 times, and in each randomized sample recalculated the RE-AN and the RE-AI similarity in one brain ROI or one CNN layer. This calculation produced two null hypothesis distributions for the two similarity values. The mean RE-AN or RE-AI similarity values (see below) were then tested against the null similarity distributions.

To examine the difference between the RE-AI and RE-AN similarity values, we resampled activity corresponding to 40 images with replacement, and in each randomized sample recalculated the difference between two similarity values in one brain ROI or one CNN layer. We then tested the mean RE-AI value against the bootstrapped distribution of RE-AN images (i.e., obtain the percentage).

*Searchlight RSA*. We also performed a surface-based searchlight analysis in order to show the cortical topology of the RE-AN and the RE-AI similarity values. For each vertex, the same 2D searchlight disk was defined as above. We then repeated the same RSA on the brain, producing two cortical maps with respect to the RE-AN and RE-AI similarity values.

**Forward encoding modeling.** We first trained the forward encoding models only based on the RE images data in the brain and the CNN. For the response sequence $\mathbf{y} = \{y_1, \text{K}, y_d\}^\text{T}$ of one vertex to the 40 RE images, it is expressed as Eq. (4):

$$\mathbf{y} = \mathbf{Xw}, \qquad (4)$$

$\mathbf{X}$ is a $m$-by-$(n+1)$ matrix, where $m$ is the number of training images (i.e., 40), and $n$ is the number of units in one CNN layer. The last column of $\mathbf{X}$ is a constant vector with all elements equal to 1. $\mathbf{w}$ is a $(n+1)$-by-1 unknown weighting matrix to solve. Because the number of training samples $m$ was less than the number of units $n$ in all CNN layers, we imposed an additional sparse constraint on the forward encoding models to avoid overfitting:

$$\min_{\mathbf{w}} \|\mathbf{w}\|_0 \quad \text{subject to} \quad \mathbf{y} = \mathbf{Xw}, \qquad (5)$$

Sparse coding has been widely suggested and used in both neuroscience and computer vision [23, 24]. We used the regularized orthogonal matching pursuit (ROMP) method to solve the sparse representation problem. ROMP is a greedy method developed by Needell D and R Vershynin [25] for sparse recovery. Features for prediction can be automatically selected to avoid overfitting. For the selected 200 vertices in each human ROI, we established 8 forward encoding models corresponding to the 8 CNN layers. This approach yielded a total of 40 forward encoding models (5 ROIs × 8 layers) for one subject.

Based on the train forward encoding models, we calculated the Pearson correlation between the empirically measured and model-predicted response patterns evoked by the adversarial images. To test the prediction accuracy against null hypotheses, we randomized the image labels and performed permutation tests as described above. Error bars (Fig. 6) indicate 95% confidence intervals obtained by bootstrapping across vertices in a brain ROI. Specifically, we resampled 80% vertices in a brain ROI 1000 times without replacement and in each sample recalculated the response prediction accuracy, resulting in a bootstrapped accuracy distribution. The upper and lower bounds of the 95% confidence intervals were derived from the bootstrapped distribution. Similarly, we compared the two bootstrapped distributions to derive the statistical difference between the RE-AI and the RE-AN similarity.

**RESULT**

*Dissociable neural representations of adversarial images in AlexNet and the human brain*

**Human brain**

For one brain ROI, we calculated the representational dissimilarity matrix (i.e., 40 x 40 RDM) for each of the three image types. We then calculated the RE-AN similarity—the correlation between the RDM of the RE images and that of the AN images, and the RE-AI similarity between the RE images and the AI images.

We made three major observations. First, the RE-AI similarities were significantly higher than null hypotheses in almost all ROIs in the three subjects (red bars in Fig. 3, permutation test, all p-values < 0.005, see Methods for the deviation of null hypotheses). Conversely, this was not true for the RE-AN similarities (blue bar in Fig. 3, permutation test, only four p-values < 0.05 in 3 subjects x 5 ROI = 15 tests). Third and more importantly, we found significantly higher RE-AI similarities than the RE-AN similarities in all ROIs (Fig. 3, bootstrap test, all p-values < 0.0001). These results suggest that the neural representations of the AI images, compared with the AN images, are much more similar to that of the corresponding RE images. Notably, this representational structure is also consistent with the perceptual similarity of the three types of images in humans. In the words, the neural representations of all images in the human brain largely echo their perceptual similarity.

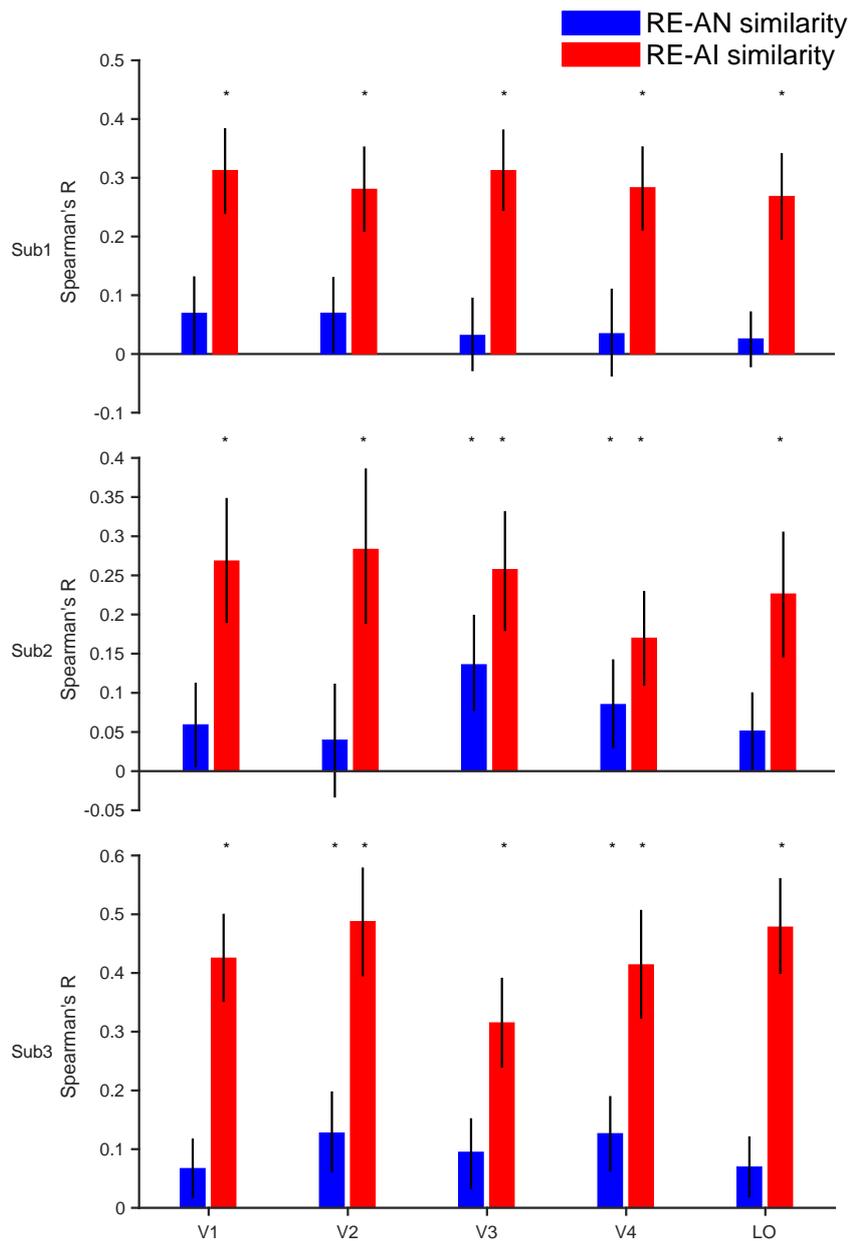

**Figure 3.** RE-AI and RE-AN similarities in the human brain. Three subplots indicate the three human subjects. In all five brain ROIs, the RE-AI (red bars) similarities are substantially higher than the RE-AN (blue bars) similarities. Error bars are 68% confidence intervals of similarity values by bootstrapping vertices in one brain ROI (see Methods). The black asterisks above bars indicate that the similarity values are significantly different from null hypotheses (permutation test, $p < 0.05$, see Methods).

We also performed a searchlight analysis to examine the cortical topology of the neural representations. The searchlight analysis used the same calculation as above (see Methods). We replicated the results (see Fig. 4) and found in a distributed pattern of higher

RE-AI similarities in early human visual cortex.

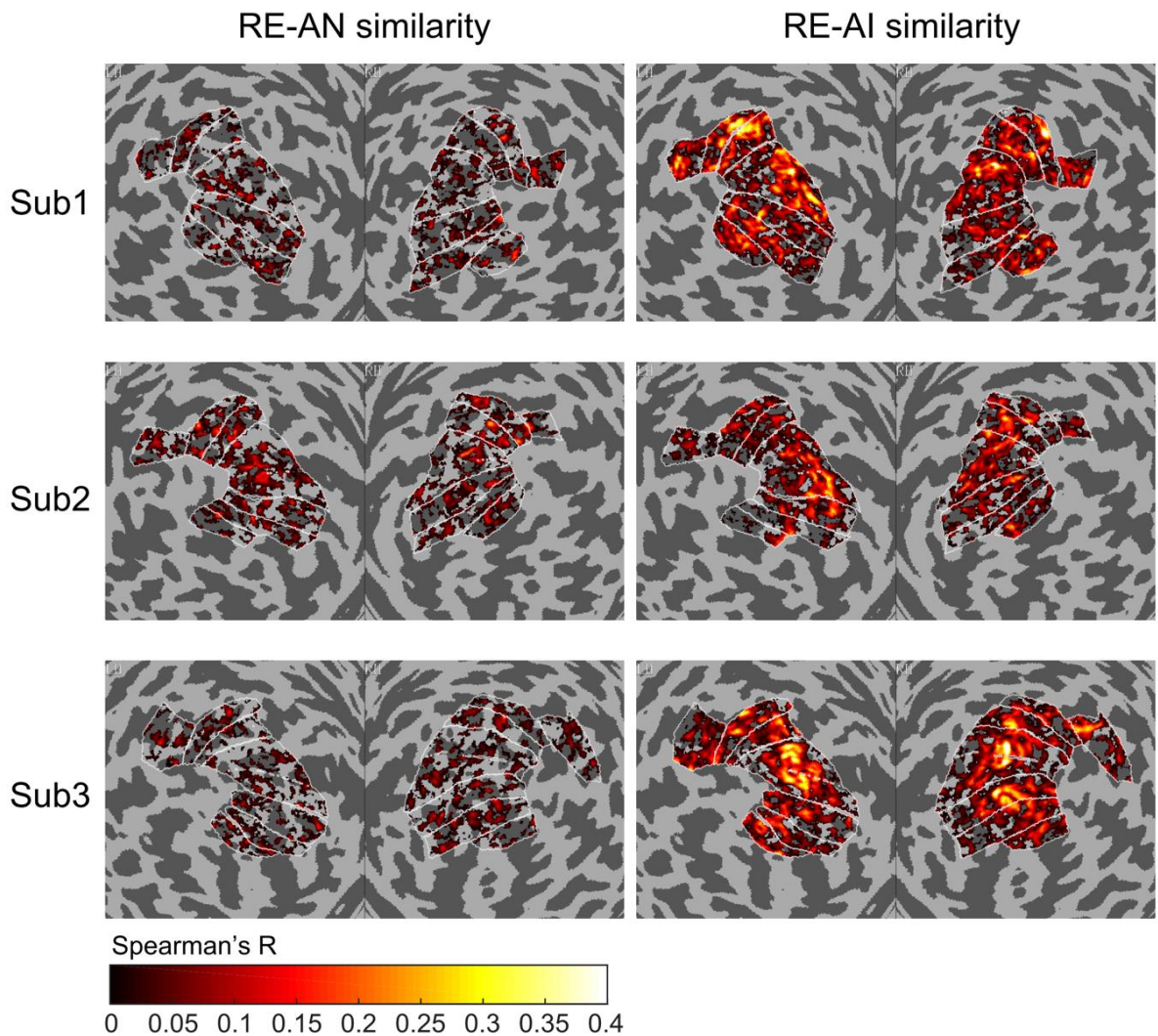

**Figure 4.** Cortical topology of RE-AI and RE-AN similarities. The RE-AI similarities are overall higher than the RE-AN similarities across all early visual areas in the human brain.

**AlexNet**

We repeated our analyses above in AlexNet and again made three observations. First, the RE-AI similarities were higher than null hypotheses across all layers (Fig. 5, permutation test, all p-values < 0.001), and the RE-AI similarities declined from low to high layers (Mann–Kendall test, p = 0.009). Second, the RE-AN similarities were initially low (p-values > 0.05 in layers 1-2) but then dramatically increased (Mann–Kendall test, p < 0.001) and became higher than the null hypotheses from layer 3 (all p-values < 0.05 in layers 3-8).

Third and most importantly, we found that the RE-AN similarities were not higher than the RE-AI similarities in all intermediate layers (i.e., layers 1-7, bootstrap test, all p-values < 0.05, layer 7, p = 0.375) except the output layer (i.e., layer 8, p < 0.05).

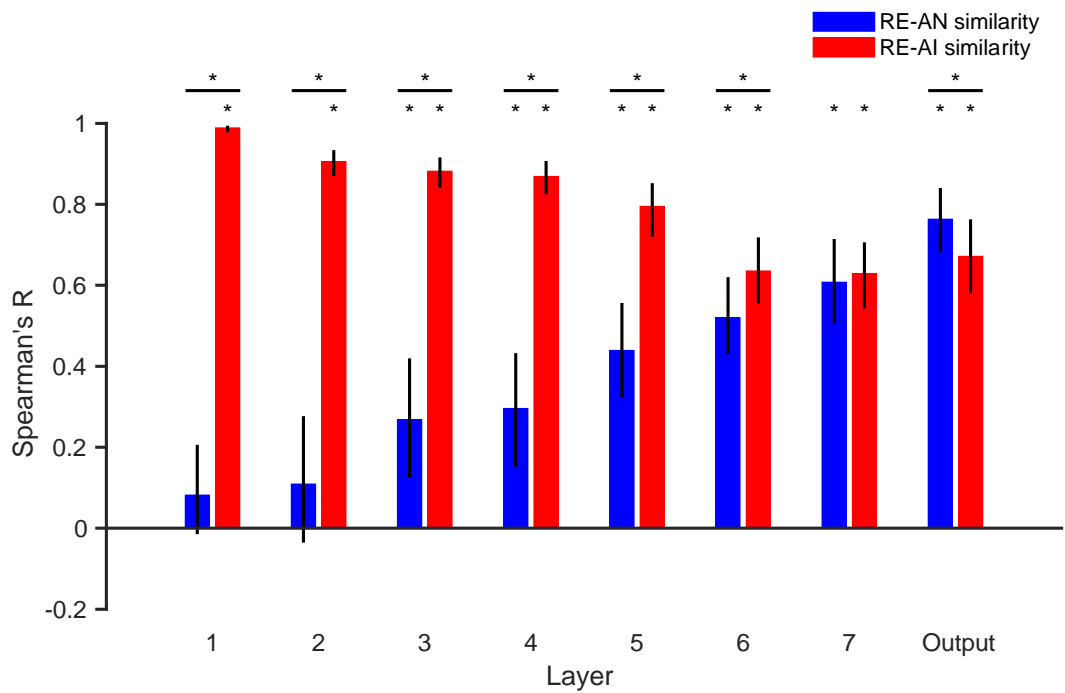

**Figure 5.** RE-AN and RE-AI similarities across layers in AlexNet. the RE-AN similarity increases and the RE-AI similarities decline along the processing hierarchy. The RE-AN similarities are not higher than the RE-AI similarities in all representational layers (i.e., layer 1-7). Error bars indicate 95% bootstrapped confidence intervals (see Methods).

These results are surprising because it suggests that neural representations of the AI images, compared with the AN images, are more similar to the representations of the RE images. However, AlexNet "perceives" the AN images more similar to corresponding RE images. In other words, there exists substantial inconsistency between the representational similarity and perceptual similarity in AlexNet. We emphasize that, assuming that in order for two images look similar, there must be at least *some* neural populations somewhere in a visual system that represent them similarly. But it is astonishing that we found no perception-compatible neural representations in *any* representational layer. Also, the transformation from layer 7 to the output layer is critical and eventually renders the RE-AN similarity higher than the RE-AI similarity in the output layer. This is idiosyncratic because AlexNet does not implement effective neural codes of objects in representational layers beforehand but the last transformation completely reverses perceptual outcomes. This is drastically different from the human brain that forms correct neural codes in all early visual areas.

*Forward encoding modeling bridges responses in AlexNet and human visual cortex*

The RSA above mainly focuses on the comparisons across image types within one visual system. We next used forward encoding modeling to directly bridge neural representations across the two systems. Forward encoding models assume that the activity of a voxel in the brain can be modeled as the linear combination of the activity of multiple artificial neurons in CNNs. Following this approach, we trained a total of 40 (5 ROIs x 8 layers) forward encoding models for one subject using regular images. We then tested how well these trained forward encoding models can generalize to the corresponding adversarial images. The rationale is that, if the brain and AlexNet process images in a similar fashion, the forward encoding models trained on the RE images should transfer to the adversarial images, and vice versa if not.

We made two major findings here. First, almost all trained encoding models successfully generalized to the AI images (Fig. 6, warm color bars, permutation test, p-values $< 0.05$ for 113 out of the 120 models for three subjects) but not to the AN images (Fig. 6, cold color bars, permutation test, p-values $> 0.05$ for 111 out of the 120 models). Second, the forward encoding models exhibited much stronger predictive power on the AI images than the AN images (bootstrap test, all p-values $< 0.05$, except the encoding model based on layer 8 for LO in subject 2, $p = 0.11$). These results suggest that the functional correspondence between AlexNet and the human brain only holds when processing RE and AI images but not AN images. This result is also consonant with the RSA above and demonstrates that both systems treat RE and AI images similarly, but AN images very differently. But again, note that AlexNet exhibits the opposite behavioral pattern of human vision.

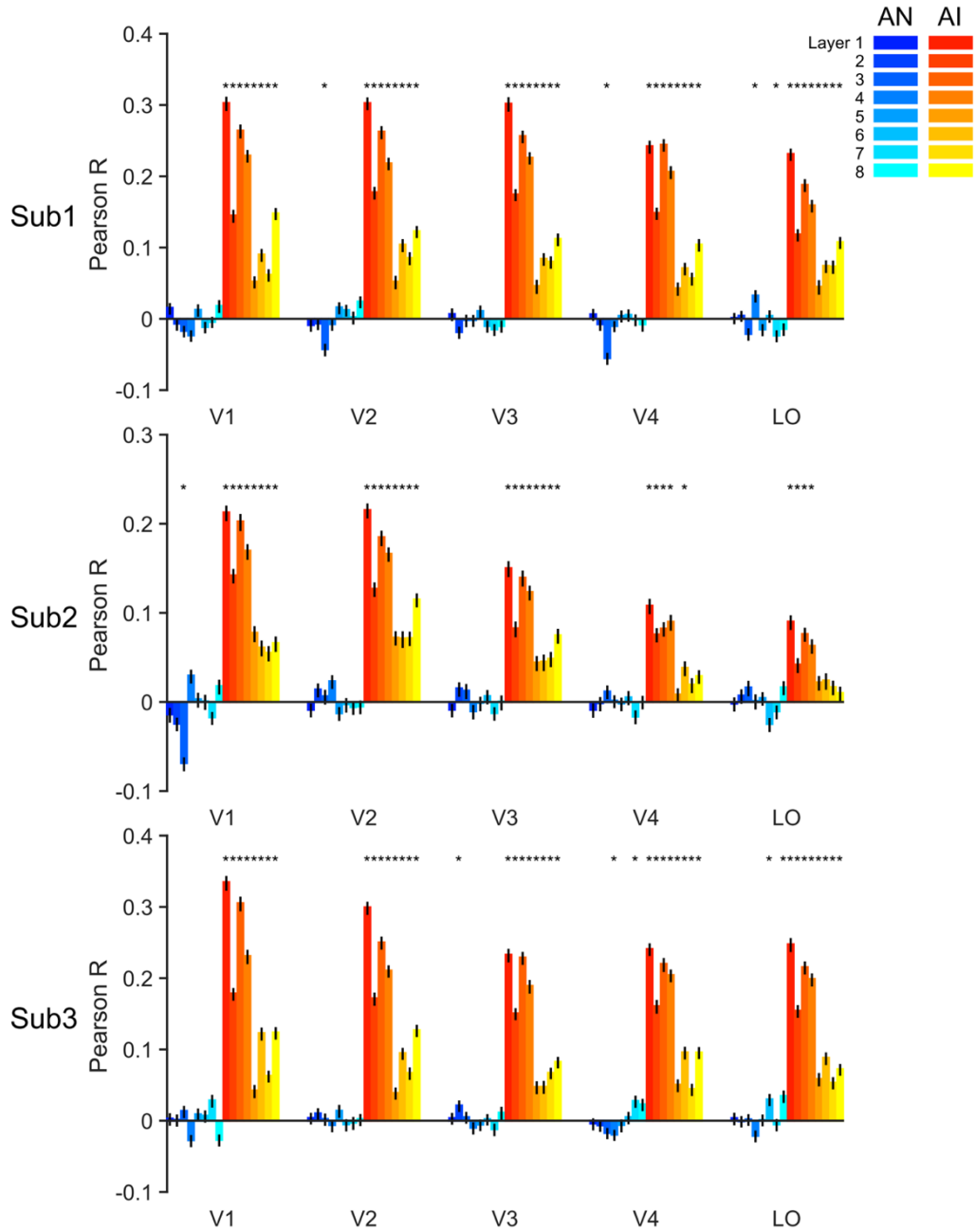

**Figure 6.** Accuracy of forward encoding models trained on RE images and then tested on adversarial images. After the models are fully trained on the RE images, we input the adversarial images as inputs to the models can predict corresponding brain responses. The y-axis indicates the Pearson correlation between the brain responses predicted by the models and the real brain responses. The generalizability of forward encoding models indicates the processing similarity between the RE and AN (cool colors) or AI (warm colors) images. Error bars indicate 95% bootstrapped confidence intervals (see Methods).

**DISCUSSION AND CONCLUSION**

Given that current CNNs still fall short in many tasks, we use adversarial images to probe the functional differences between a prototypical CNN—AlexNet, and the human visual system. We make three major findings. First, the representations of AI images, compared with AN images, are more similar to the representations of corresponding RE images. These representational patterns in the brain are consistent with human percepts (i.e., perceptual similarity). Second, we discover a representation-perception disassociation in *all* intermediate layers in AlexNet. Third, we use forward encoding modeling to link neural activity in both systems. Results show that the processing of RE and AI images are quite similar but both are significantly different from AN images. Overall, these observations demonstrate the capacity and limit of the similarities between current CNNs and human vision.

*Abnormal neural representations of adversarial images in CNNs*

To what extent neural representations reflect physical or perceived properties of stimuli is a key question in modern vision science. In the human brain, researchers have found that early visual processing mainly processes low-level physical properties of stimuli, and late visual processing mainly supports high-level categorical perception [26]. We ask a similar question here—to what extent neural representations in CNNs or the human brain reflect their conscious perception.

One might argue that the representation-perception disassociation in AlexNet is trivial, given that we already know that AlexNet exhibits opposite behavioral patterns compared to human vision. But we believe thorough quantifications of their neural representations in both systems are still of great value. First, neural representations do not necessarily follow our conscious perception, and numerous neuroscience studies have shown disassociated neural activity and perception in both the primate or human brain in many cases, such as visual illusion, binocular rivalry, visual masking [27]. The question of representation-perception association lies at the center of the neuroscience of consciousness and should also be explicitly addressed in AI research. Second, whether representation and perception are consistent or not highly depends on processing hierarchy, which again needs to be carefully quantified across visual areas in the human brain and layers in CNNs. Here, we found no similar representations of AN and regular images in any intermediate layer in AlexNet even though they "look" similar. This is analogous to the scenario that we cannot decode any similar representational patterns of two images throughout a subject's brain, although the subject behaviorally reports the two images are similar.

*Adversarial images as a tool to probe functional differences between the CNN and human vision*

In computer vision, adversarial images impose problems on the real-life applications of artificial systems (i.e., adversarial attack) [28]. Several theories have been proposed to explain the phenomenon of adversarial images [29]. For example, one possible explanation is that CNNs are forced to behave linearly in high dimensional spaces, rendering them vulnerable to adversarial attacks [30]. Besides, flatness [31] and large local curvature of the decision boundaries [32], as well as low flexibility of the networks [33] are all possible reasons. Szegedy C et al. [14] has suggested that current CNNs are essentially complex nonlinear classifiers, and this discriminative modeling approach does not consider generative distributions of data. We will further address this issue in the next section.

In this study, we focused on one particular utility of adversarial images—to test the dissimilarities between CNNs and the human brain. Note that although the effects of adversarial images indicate the deficiencies of current CNNs, we do not object the approach to use CNNs as a reference to understand the mechanisms of the brain. Our study here fits the broad interests in comparing CNNs and the human brain in various aspects. We differ from other studies just because we focus on their differences. We do acknowledge that it is quite valuable to demonstrate functional similarities between the two systems. But we believe that revealing their differences, as an alternative approach, might further foster our understandings of how to improve the design of CNNs. This is similar to the logic of using ideal observer analysis in vision science. Although we know human visual behavior is not optimal in many situations, the comparison to an ideal observer is still meaningful as it can reveal some critical mechanisms of human visual processing.

Some recent efforts have been devoted to addressing CNN-human differences. For example, Rajalingham et al. [34] found that CNNs explain human (or non-human primate) rapid object recognition behavior at the level of category but not individual images. CNNs better explains the ventral stream than the dorsal stream [35]. To further examine their differences, people have created some unnatural stimuli/tasks, and our work on adversarial images follows this line of research. The rationale is that, if CNNs are similar to humans, they should exhibit the same capability in both ordinary and unnatural circumstances. A few studies adopted some other manipulations [36, 37], such as manipulation of image noise [38] and distortion [39].

*Possible caveats of CNNs in the processing of adversarial images*

Why CNNs and human vision behave differently on adversarial images, especially on AN images? We want to highlight three reasons and discuss the potential route to circumvent them.

First, current CNNs are trained to match the classification labels generated by humans. This approach is a discriminative modeling approach that characterizes the probability of *p*(class | image). Note that natural images only occupy a low-dimensional manifold in the entire image space. Under this framework, there must exist a set of artificial images in the image space that fulfill a classifier but do not belong to any distribution of real images. Humans cannot recognize AN images because humans do not merely rely on discriminative classifiers but instead perform Bayesian inference and take into consideration both likelihood *p*(image|class) and prior experience *p*(class). One approach to overcome this is to build generative deep models to learn latent distributions of images, such as variational autoencoders [40] and generative adversarial networks [41].

Another advantage of deep generative models is to explicitly model the uncertainty in sensory processing and decision. It has been well-established in cognitive neuroscience that the human brain computes not only form a categorical perceptual decision, but also a full posterior distribution over all possible hidden causes given a visual input [42-44]. This posterior distribution is also propagated to downstream decision units and influences other aspects of behavior.

Third, more recurrent and feedback connections are needed. Numerous studies have shown the critical role of top-down processing in a wide range of visual tasks, including recognition [45, 46], tracking [47], as well as other cognitive domains, such as memory [48], language comprehension [49] and decision making [50, 51]. In our results, the responses in human visual cortex likely reflect the combination of feedforward and feedback effects whereas the activity in most CNNs only reflects feedforward inputs from earlier layers. A recent study has shown that recurrence is necessary to predict neural dynamics in the human brain using CNN features [52].

*Concluding remarks*

In the present study, we compared neural representations of adversarial images in AlexNet and the human visual system. Using RSA and forward encoding modeling, we found that the neural representations of RE and AI images are similar in both systems but AN images were idiosyncratically processed in AlexNet. These findings open a new avenue to help design CNN architectures to achieve brain-like computation.

**AUTHOR CONTRIBUTION**

C.Z., R-Y.Z., and L.T. designed the research. C.Z., and L.T. collected the data. C.Z., and R-Y.Z. analyzed the data and wrote the manuscript.

**ACKNOWLEDGMENTS**

We thank Pinglei Bao, Feitong Yang, Baolin Liu, and Huafu Chen for their invaluable comments on the manuscript. This work was supported by the National Key Research and Development Plan of China under grant 2017YFB1002502, the National Natural Science Foundation of China (No.61701089), and the Natural Science Foundation of Henan Province of China (No.162300410333).

**DISCLOSURE OF POTENTIAL CONFLICT OF INTEREST**

The authors declare that they have no competing interests. All procedures followed were in accordance with the ethical standards of the responsible committee on human experimentation (Henan Provincial People's Hospital) and with the Helsinki Declaration of 1975, as revised in 2008 (5). Informed consent was obtained from all patients for being included in the study.

Supplementary Information for

**Dissociable neural representations of adversarially perturbed images in convolutional neural networks and the human brain**

Chi Zhang, Xiaohan Duan, Linyuan Wang, Yongli Li, Bin Yan, Guoen Hu, Ru-Yuan Zhang, Li Tong

**Table of Content**

**Supplementary Note 1: Hierarchical correspondence between AlexNet and human visual cortex only holds for AI images**
**Figure S1. Percentage of vertices explained by encoding models in each ROI**
**Figure S2. RE-AN and RE-AI similarities with 400 vertices in each ROI**

**Supplementary Note 1: Hierarchical correspondence between AlexNet and human visual cortex only holds for AI images**

Previous studies have revealed that the tuning complexity along the human ventral stream corresponds to the feature complexity from low- to high-level layers in CNNs [1, 2]. We thus investigated the hierarchical correspondence between AlexNet and the human brain under different image conditions. In each brain ROI, we calculated the proportion of vertices whose responses can be best explained by the units in each layer in AlexNet. This yielded a 1 x 8 vector of the proportion values for that brain ROI. This calculation was performed independently in each subject and in each of the two presentation trials. We could thus obtain six independent measurements (3 subjects x 2 trials) for the proportion vector in this ROI. We then averaged the proportion values of the low-level layers (i.e., layers 1&2) across subjects and trials, and also high-level layers (i.e., layers 7&8). A paired t-test was performed to examine the difference between the proportion values of the low-level and the high-level layers.

Since all brain ROIs here are early visual areas, we expected that the forward encoding models using low-level features should better characterize empirically measured cortical responses. Indeed, this pattern was observed in the AI images (Fig. S1)–that is, the low-level CNN layers, compared with the high-level CNN layers, better predict the vertex responses (layers 1-2 vs. 7-8, paired t-test, $p < 0.0001$) across almost all ROIs. On the AN images, however, no significant difference was detected between the predictive power of the low-level and the high-level layers (layers 1-2 vs. 7-8, paired t-test, $p = 0.2351$). Moreover, we found that, from V1 to LO, the higher-level CNN layers predicted more and more vertices' responses (V1 vs. LO, paired t-test, $p = 0.0004$) towards the AI images. But we did not observe such trend in the AN images (V1 vs. LO, paired t-test, $p = 0.5631$). In other words, the hierarchical correspondence between AlexNet and the human brain observed in regular images also holds in AI images but not in AN images, further suggesting the idiosyncratic processing of AN images in AlexNet.

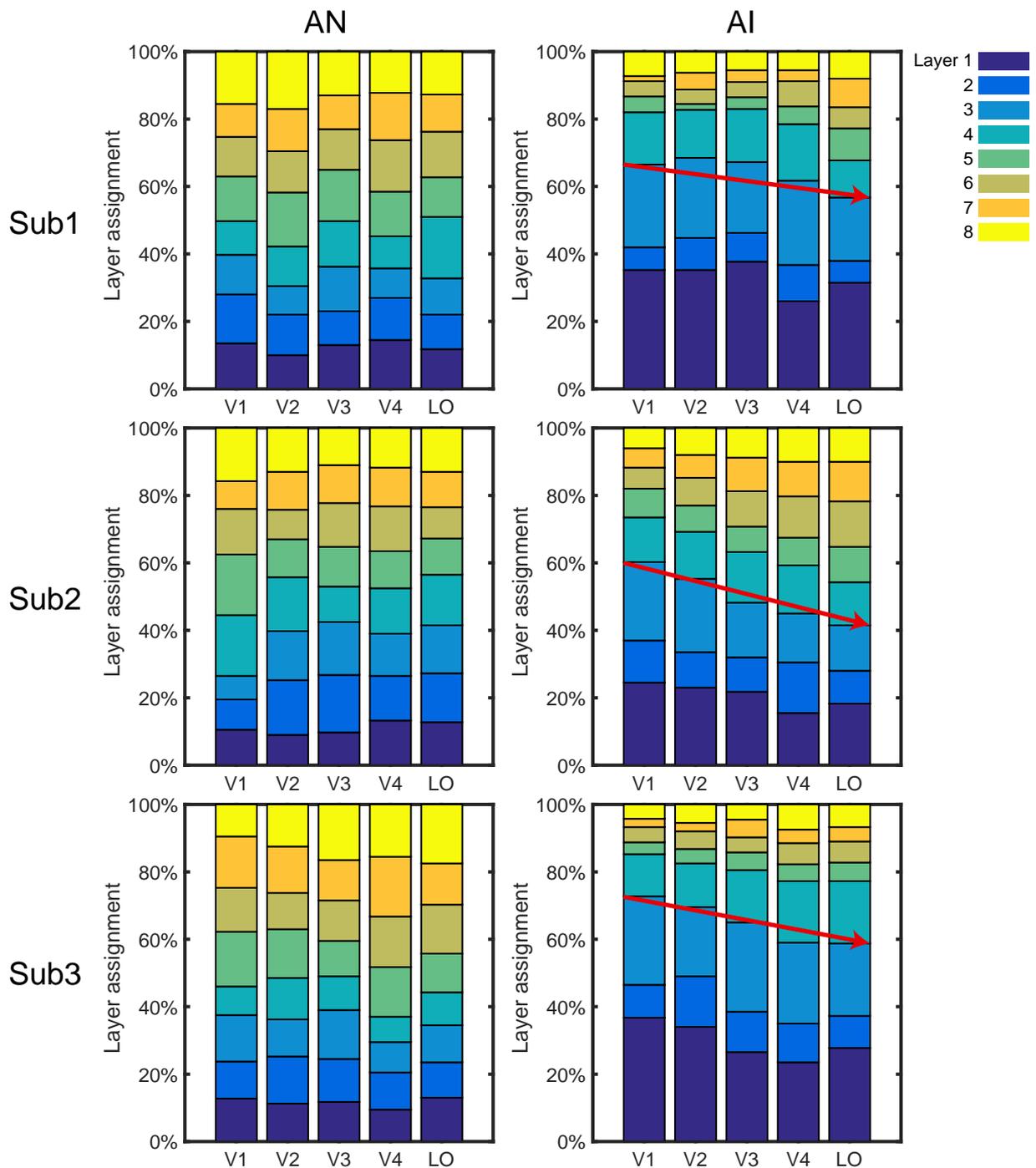

**Figure S1.** The percentage of vertices that can be best explained by features in each CNN layer. For the AI images, we replicate the hierarchical correspondence between AlexNet and the human brain—early visual areas can be better explained by feature activity in low-level CNN layers. This pattern is, however, not obvious for the AN images. Moreover, the proportion of vertices assigned to high-level CNN layers decreases along processing hierarchy (i.e., V1 to LO) for the AI images (indicated by the red arrows), but not for the AN images. These results indicate that the hierarchical correspondence between the AlexNet and the human brain only holds for the AI images.

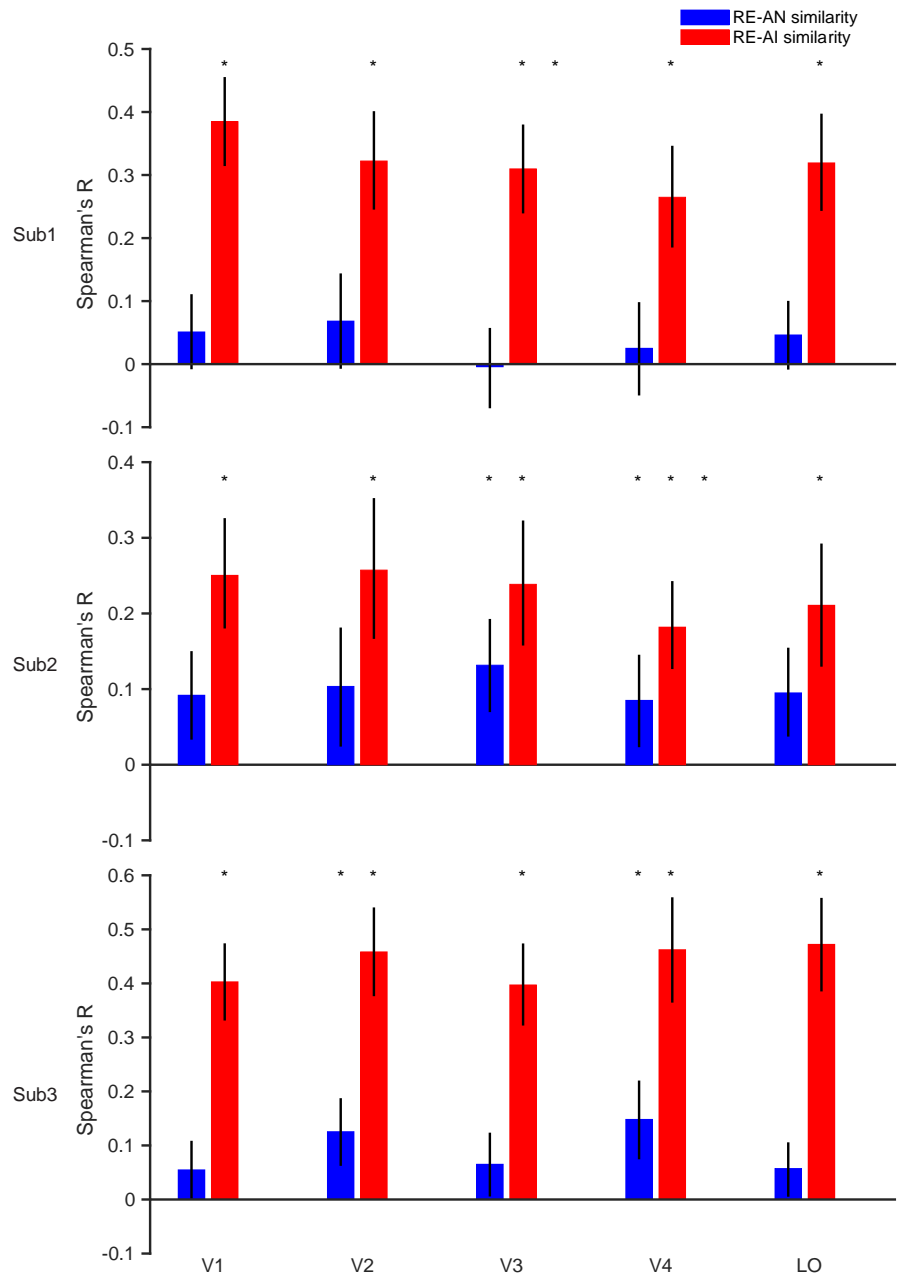

Fig. S2. RE-AI and RE-AN similarity in the human brain. This figure is similar to Fig. 3 except that 400 vertices are selected in each brain region.